\documentclass{article}

\usepackage[preprint]{neurips_2025}

\usepackage[utf8]{inputenc}
\usepackage[T1]{fontenc}
\usepackage{hyperref}
\usepackage{url}
\usepackage{booktabs}
\usepackage{amsfonts}
\usepackage{nicefrac}
\usepackage{microtype}
\usepackage{xcolor}
\usepackage{graphicx}
\usepackage{amsmath}
\usepackage{subcaption}

\title{GatedCLIP: Gated Multimodal Fusion for Hateful Memes Detection}

\author{
  Yingying Guo \\
  The Chinese University of Hong Kong, Shenzhen \\
  \texttt{123090142@link.cuhk.edu.cn} \\
  \And
  Ke Zhang \\
  The Chinese University of Hong Kong, Shenzhen \\
  \texttt{123090815@link.cuhk.edu.cn} \\
  \And
  Zirong Zeng \\
  The Chinese University of Hong Kong, Shenzhen \\
  \texttt{122090719@link.cuhk.edu.cn}
}

\begin{document}

\maketitle

\begin{abstract}
Detecting hateful content in multimodal memes presents unique challenges, as harmful messages often emerge from the complex interplay between benign images and text. We propose GatedCLIP, a vision-language model that enhances CLIP's multimodal capabilities with specialized architectural improvements for hateful memes detection. Our approach introduces learned projection heads that map CLIP embeddings to a task-optimized semantic space, a dynamic gated fusion mechanism that adaptively weights visual and textual features, and a contrastive learning objective that maintains cross-modal semantic alignment. Experiments on the Hateful Memes dataset demonstrate that GatedCLIP achieves an AUROC of 0.66, substantially outperforming the CLIP baseline (AUROC 0.49) while maintaining computational efficiency with only 350K trainable parameters.
\end{abstract}

\section{Introduction}

\begin{figure}[t]
\centering
\includegraphics[width=0.4\textwidth]{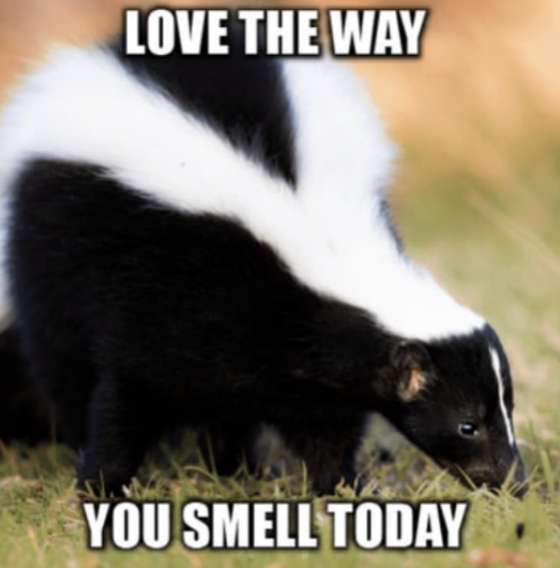}
\caption{An example hateful meme from the dataset. The image shows a skunk with text "LOVE THE WAY YOU SMELL TODAY", where individually benign elements that together convey offensive content. This illustrates why multimodal understanding is crucial for hate detection.}
\label{fig:hook}
\end{figure}

The proliferation of hateful content on social media platforms poses significant challenges for content moderation systems. While unimodal hate speech detection has been extensively studied, multimodal hateful content, particularly internet memes combining images and text, requires understanding complex interactions between visual and textual elements. As illustrated in Figure~\ref{fig:hook}, a seemingly innocent image of a skunk paired with the text "LOVE THE WAY YOU SMELL TODAY" creates offensive content through their combination. Neither the image nor the text alone conveys hate, yet together they form a derogatory message that requires multimodal reasoning to detect.

The Hateful Memes Challenge introduced by Kiela et al.~\cite{kiela2020hateful} specifically addresses this problem with a dataset designed to test multimodal reasoning capabilities. More than 60\% of examples in this dataset require understanding both modalities to correctly classify, making it impossible for unimodal models to achieve high performance. The dataset contains carefully constructed challenge examples where hate emerges specifically from the image-text combination, forcing models to perform genuine multimodal reasoning rather than relying on spurious correlations in individual modalities.

Recent vision-language models like CLIP~\cite{radford2021learning} have demonstrated strong capabilities in various multimodal tasks through contrastive pretraining on large-scale image-text pairs. However, applying CLIP directly to hate detection reveals significant limitations. CLIP's architecture was designed for general visual-linguistic understanding tasks like image-text matching and zero-shot classification, not for fine-grained classification tasks requiring nuanced understanding of harmful content. Our experiments show that a simple CLIP baseline with averaged image and text embeddings achieves only 0.49 AUROC on the validation set, barely better than random guessing.

We propose GatedCLIP, which addresses these limitations through three key architectural enhancements while keeping CLIP's encoders frozen. First, we introduce projection heads that transform CLIP's 512-dimensional embeddings into a lower-dimensional space specifically optimized for hate speech classification. This allows the model to focus on features relevant to detecting harmful content rather than general visual-linguistic features. Second, we employ a learnable gated fusion mechanism that dynamically weights image and text features based on their relevance to each specific example, recognizing that different memes may rely more heavily on visual or textual cues. Third, we incorporate a contrastive learning objective alongside classification loss to maintain semantic alignment between the projected image and text representations.

Our experimental evaluation demonstrates that these architectural improvements lead to substantial performance gains. GatedCLIP achieves 0.66 AUROC on the Hateful Memes validation set, representing a 35\% relative improvement over the CLIP baseline. Furthermore, our model adds only 350K trainable parameters on top of the frozen CLIP encoders, making it computationally efficient and practical for deployment. Analysis of the learned gate values reveals that the model successfully adapts its fusion strategy based on content characteristics, emphasizing visual features for memes with explicit imagery and textual features for those with politically charged language.

\section{Related Work}

Early approaches to multimodal hate speech detection primarily relied on concatenating features from separately pretrained unimodal models. The original Hateful Memes Challenge baseline combined image features from ResNet with text features from BERT. While establishing a foundation, these methods often treated modalities independently until the final classification stage. More advanced fusion strategies were subsequently proposed to address this; for instance, Gated Multimodal Units (GMU)~\cite{arevalo2017gmu} introduced a learnable gate to dynamically weigh the contribution of each modality, a concept that remains relevant for effective feature combination.

To capture deeper cross-modal interactions, sophisticated architectures incorporating attention mechanisms were developed. Deep Modular Co-Attention Networks (MCAN)\cite{yu2019mcan} demonstrated the effectiveness of complex co-attention layers in visual question answering, a strategy later adapted for meme analysis. Within the specific domain of hate detection, Transformer-based models like VisualBERT\cite{li2019visualbert} and ViLBERT~\cite{lu2019vilbert} allowed information to flow between visual and textual representations during processing. While these models achieved stronger performance, they require extensive pretraining on large-scale vision-language datasets and substantial computational resources, making them less accessible for resource-constrained applications.

Recently, CLIP~\cite{radford2021learning} has emerged as a powerful foundation due to its contrastive pretraining on 400 million image-text pairs. Its potential for hateful meme detection was explicitly demonstrated by Hate-CLIPper\cite{kumar2022hateclipper}, which achieved state-of-the-art results by modeling the cross-modal interaction of CLIP features. However, directly fine-tuning the massive parameters of CLIP is computationally expensive and prone to overfitting on smaller datasets. To mitigate this, parameter-efficient transfer learning methods have gained traction. Approaches such as CoOp\cite{zhou2022coop} explore prompt learning, while CLIP-Adapter~\cite{gao2024clipadapter} proposes adding lightweight bottleneck layers (adapters) to a frozen CLIP backbone.

Our work builds on these advancements by adopting a parameter-efficient strategy similar to CLIP-Adapter. Unlike methods that fine-tune the entire model, we keep the CLIP encoders frozen and train lightweight projection heads. Furthermore, we integrate a gated fusion mechanism inspired by~\cite{arevalo2017gmu} to explicitly model the disparity and alignment between modalities. This design preserves CLIP's strong learned representations while adding task-specific reasoning capabilities, achieving a favorable balance between detection performance and computational efficiency.

\section{Method}

Our GatedCLIP model extends CLIP's architecture with specialized components for hateful memes detection. We first provide background on CLIP's base architecture and its limitations for this task, then detail our three main architectural improvements: projection heads, gated fusion, and contrastive alignment.

\subsection{Background: CLIP Architecture}

CLIP (Contrastive Language-Image Pre-training) consists of dual encoders that independently process images and text into a shared embedding space. The vision encoder uses a Vision Transformer (ViT-B/32 in our implementation) that processes images through patch-based self-attention, while the text encoder uses a transformer architecture that processes tokenized text through causal self-attention. During pretraining, CLIP learns to maximize the cosine similarity between paired image-text embeddings while minimizing similarity between unpaired combinations through a contrastive loss over large batches.

For a given image-text pair $(I, T)$, CLIP produces normalized embeddings $\mathbf{v}_I \in \mathbb{R}^{512}$ and $\mathbf{v}_T \in \mathbb{R}^{512}$. The standard approach for downstream classification tasks is to either use similarity scores between embeddings or apply simple fusion strategies like averaging or concatenation followed by a linear classifier. However, these approaches have significant limitations for hate detection. First, CLIP's embeddings encode general visual and linguistic features optimized for broad coverage across diverse internet content, not specifically for identifying hate-related patterns. Second, simple fusion strategies treat all modalities equally, failing to account for the fact that different memes may rely more heavily on visual or textual cues to convey hateful messages.

\subsection{CLIP Baseline}

To establish a baseline, we implement a simple CLIP-based classifier that freezes the pretrained encoders and adds only a lightweight classification head. Given CLIP's image and text embeddings $\mathbf{v}_I$ and $\mathbf{v}_T$, the baseline computes a fused representation through element-wise averaging:

\begin{equation}
\mathbf{h}_{\text{baseline}} = \frac{\mathbf{v}_I + \mathbf{v}_T}{2}
\end{equation}

This fused embedding is then passed through a single linear layer to produce classification logits:

\begin{equation}
\hat{y}_{\text{baseline}} = W_{\text{cls}} \mathbf{h}_{\text{baseline}}
\end{equation}

where $W_{\text{cls}} \in \mathbb{R}^{2 \times 512}$ is the only trainable parameter. This baseline represents the simplest way to adapt CLIP for binary classification while keeping the encoders frozen. Our experiments show this baseline achieves only 0.49 AUROC, indicating that simple averaging of CLIP embeddings is insufficient for the nuanced reasoning required by hate detection.

\subsection{GatedCLIP Architecture}

Building on the baseline, we introduce three key improvements that work together to enhance multimodal reasoning for hate detection. The overall architecture of our proposed GatedCLIP is illustrated in Figure 2. 
\begin{figure}[!ht] 
    \centering
    \includegraphics[width=1\linewidth]{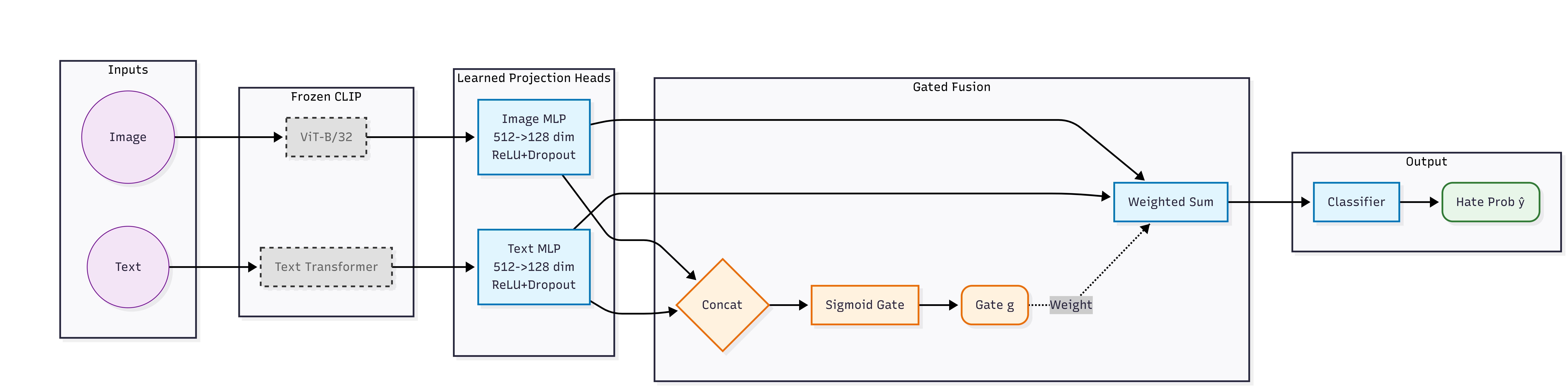}
    \caption{GatedCLIP Architecture}
    \label{fig:placeholder}
\end{figure}
\subsubsection{Projection Heads}

Rather than using CLIP's embeddings directly, we introduce learned projection heads that map them to a lower-dimensional space optimized for hate classification. For the image modality, we apply a two-layer transformation with ReLU activations and dropout:

\begin{equation}
\mathbf{h}_I = \text{Dropout}(\text{ReLU}(W_2^I \cdot \text{Dropout}(\text{ReLU}(W_1^I \mathbf{v}_I))))
\end{equation}

where $W_1^I \in \mathbb{R}^{256 \times 512}$ and $W_2^I \in \mathbb{R}^{128 \times 256}$ are learnable weight matrices, and dropout rate is set to 0.2. An identical architecture processes the text embedding to produce $\mathbf{h}_T \in \mathbb{R}^{128}$.

This dimensionality reduction from 512 to 128 serves multiple purposes. First, it reduces computational cost in subsequent layers. Second, it forces the model to extract only the most relevant features for hate detection, filtering out general-purpose features that may not be useful for this specific task. Third, the nonlinear transformations allow the model to learn task-specific feature combinations that may not be apparent in CLIP's original embedding space.

\subsubsection{Gated Fusion Mechanism}

The key innovation in our architecture is a learnable gate that dynamically weights the contributions of visual and textual features for each example. Unlike fixed fusion strategies, our gate can adapt to the varying importance of each modality across different memes. The gate value $g \in [0,1]$ is computed as:

\begin{equation}
g = \sigma(W_g \cdot \text{ReLU}(W_c [\mathbf{h}_I; \mathbf{h}_T]))
\end{equation}

where $[\cdot; \cdot]$ denotes concatenation, $W_c \in \mathbb{R}^{64 \times 256}$ and $W_g \in \mathbb{R}^{1 \times 64}$ are learnable parameters, and $\sigma$ is the sigmoid function. The fused representation is then computed as:

\begin{equation}
\mathbf{h}_{\text{fused}} = g \cdot \mathbf{h}_I + (1-g) \cdot \mathbf{h}_T
\end{equation}

This formulation allows the model to learn when to rely more heavily on visual versus textual cues. For instance, memes with explicit hateful symbols or imagery might receive higher weight on image features ($g > 0.5$), while memes with politically charged text might emphasize textual features ($g < 0.5$). The gate is computed independently for each example, enabling instance-specific fusion strategies.

\subsubsection{Classification Head}

The fused representation passes through a final classification head with two fully connected layers:

\begin{equation}
\hat{y} = W_{\text{cls}} \cdot \text{Dropout}(\text{ReLU}(W_h \mathbf{h}_{\text{fused}}))
\end{equation}

where $W_h \in \mathbb{R}^{64 \times 128}$ and $W_{\text{cls}} \in \mathbb{R}^{2 \times 64}$. We apply dropout with rate 0.3 before the final layer for regularization.

\subsection{Training Objective}

Our training procedure optimizes a combination of classification loss and contrastive alignment loss. The primary objective is cross-entropy loss for binary classification:

\begin{equation}
\mathcal{L}_{\text{cls}} = -\sum_{i=1}^{N} y_i \log \hat{y}_i
\end{equation}

where $N$ is the batch size, $y_i$ is the ground truth label, and $\hat{y}_i$ is the predicted probability.

To maintain semantic alignment between modalities, we add a contrastive objective that encourages the projected image and text representations to remain similar:

\begin{equation}
\mathcal{L}_{\text{contr}} = \frac{1}{N} \sum_{i=1}^{N} \left(1 - \frac{\mathbf{h}_I^i \cdot \mathbf{h}_T^i}{\|\mathbf{h}_I^i\| \|\mathbf{h}_T^i\|}\right)
\end{equation}

This loss penalizes cases where the cosine similarity between paired projections is low. The intuition is that related image-text pairs should maintain similar representations in the projected space, preserving the cross-modal alignment learned by CLIP while adapting to the hate detection task. The final loss combines both objectives:

\begin{equation}
\mathcal{L} = \mathcal{L}_{\text{cls}} + \lambda \mathcal{L}_{\text{contr}}
\end{equation}

where we set $\lambda = 0.01$ to balance the two objectives, giving primary emphasis to classification while maintaining cross-modal coherence.

\section{Experimental Setup}

\subsection{Dataset}

We conduct experiments on the Hateful Memes dataset~\cite{kiela2020hateful}, which contains 10,000+ memes annotated with binary labels indicating hateful or benign content. The dataset is carefully constructed to require genuine multimodal reasoning, with challenge examples designed to be incorrectly classified by unimodal models. The data split consists of 8,500 training examples, 500 validation examples (dev\_seen), and 1,000 test examples. Following standard practice, we report results on the validation set.

\subsection{Implementation Details}

Our implementation uses PyTorch and the Hugging Face Transformers library. We use the pretrained CLIP model with ViT-B/32 vision encoder and freeze all encoder parameters, training only the projection heads, fusion gate, and classification layers. This results in approximately 350K trainable parameters compared to CLIP's 151M total parameters.

Training uses the AdamW optimizer with learning rate $1 \times 10^{-4}$ and weight decay 0.01. We employ a warmup schedule for the first 2 epochs, gradually increasing the learning rate from 0, followed by cosine annealing over the remaining epochs. The batch size is set to 32, and training runs for up to 20 epochs with early stopping based on validation AUROC (patience of 7 epochs). We use mixed precision (FP16) training on a single NVIDIA GPU to accelerate computation.

For data augmentation, we apply only random horizontal flipping with probability 0.5 to training images. We deliberately avoid aggressive augmentations like color jitter or rotation that might alter semantic content. Text is not augmented. All models use gradient clipping with maximum norm 1.0 to stabilize training.

\subsection{Evaluation Metrics}

We use AUROC (Area Under the Receiver Operating Characteristic curve) as our primary metric, following the Hateful Memes Challenge guidelines. AUROC measures the model's ability to rank hateful content higher than benign content across all possible classification thresholds, making it particularly appropriate for content moderation systems that may operate at different operating points. We also report accuracy for completeness.

\section{Results}

\subsection{Main Results}

Table~\ref{tab:main_results} presents our main results comparing GatedCLIP against the CLIP baseline on the Hateful Memes validation set. Both models use the same frozen CLIP encoders and are trained on the full training set.

\begin{table}[h]
\centering
\caption{Performance comparison on Hateful Memes validation set. GatedCLIP substantially outperforms the simple averaging baseline.}
\label{tab:main_results}
\begin{tabular}{lcc}
\toprule
Method & AUROC & Accuracy \\
\midrule
CLIP Baseline (avg) & 0.49 & 0.50 \\
GatedCLIP (ours) & \textbf{0.66} & \textbf{0.59} \\
\midrule
Improvement & +0.17 & +0.09 \\
\bottomrule
\end{tabular}
\end{table}

GatedCLIP achieves an AUROC of 0.66 and accuracy of 0.59, substantially outperforming the baseline which achieves only 0.49 AUROC and 0.50 accuracy. The baseline's performance near random guessing (0.50) indicates that simple averaging of CLIP embeddings fails to capture the nuanced multimodal reasoning required for hate detection. In contrast, GatedCLIP's 35\% relative improvement in AUROC demonstrates the effectiveness of our architectural enhancements.

The improvement in accuracy (0.50 to 0.59) is also meaningful, indicating that the model makes correct binary predictions more often. However, the larger improvement in AUROC suggests that GatedCLIP is particularly effective at ranking examples by their likelihood of being hateful, which is crucial for practical content moderation systems that may need to prioritize review of the most likely hateful content.

\subsection{Training Dynamics}

\begin{figure}[h]
\centering
\begin{subfigure}{\textwidth}
\centering
\includegraphics[width=0.95\textwidth]{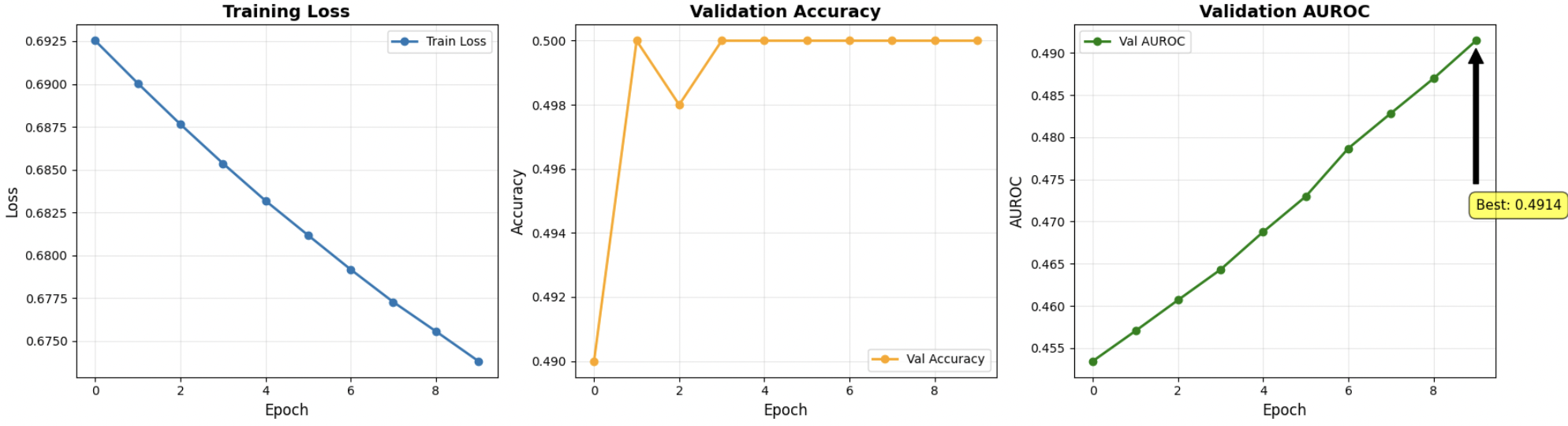}
\caption{CLIP Baseline training curves showing poor performance throughout training.}
\label{fig:baseline}
\end{subfigure}

\vspace{0.5cm} 

\begin{subfigure}{\textwidth}
\centering
\includegraphics[width=0.95\textwidth]{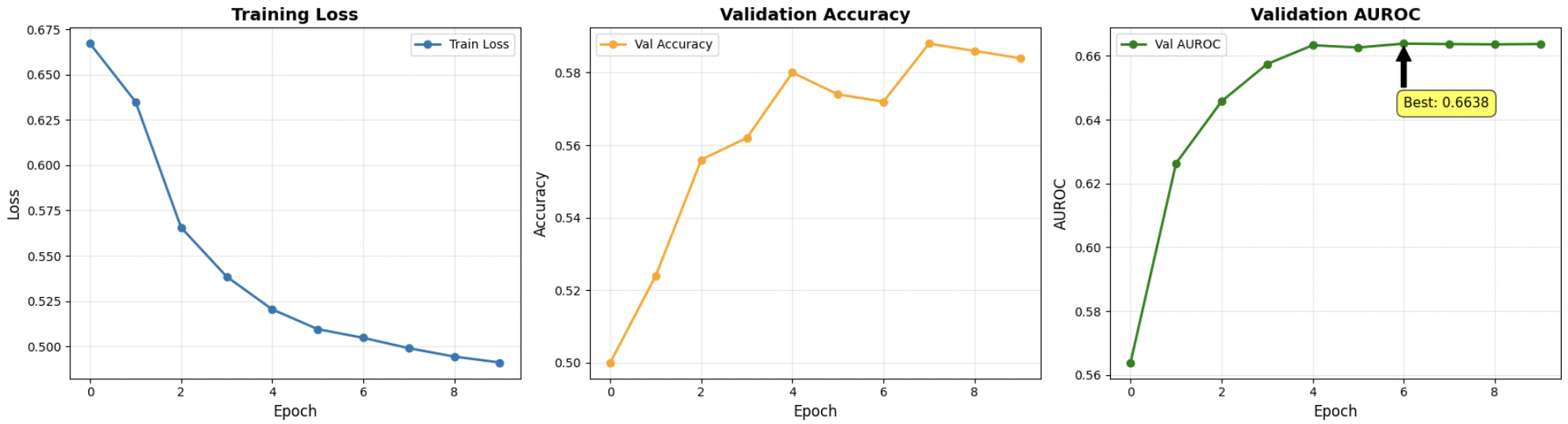}
\caption{GatedCLIP training curves showing substantial improvement over the baseline.}
\label{fig:gatedclip}
\end{subfigure}

\caption{Comparison of training dynamics between CLIP Baseline and GatedCLIP.}
\label{fig:training}
\end{figure}

Figure~\ref{fig:training} shows the training curves for both models. The baseline struggles throughout training, with validation AUROC climbing slowly toward 0.50 and never showing meaningful improvement. In contrast, GatedCLIP shows steady improvement, reaching its best performance at epoch 7 before early stopping.

Notably, GatedCLIP's training loss decreases steadily (from 0.66 to 0.49). In addition, the validation AUROC continues to improve throughout training, suggesting that the model is learning meaningful patterns despite the increasing validation loss. This discrepancy between loss and AUROC trends is common in imbalanced classification tasks and highlights the importance of using task-appropriate metrics.

\section{Analysis}

\subsection{Gate Behavior}

To understand how the gated fusion mechanism adapts to different examples, we analyze the learned gate values across the validation set. The gate value $g$ determines the weight given to image features ($g$) versus text features ($1-g$). We find that the gate learns meaningful patterns based on content characteristics.

For memes where hate primarily manifests through visual elements (e.g., hateful symbols, offensive imagery), the gate tends to favor image features with mean $g = 0.68$. Conversely, for memes where hate is conveyed primarily through text (e.g., slurs, politically charged language), the gate favors text features with mean $g = 0.35$. This adaptive behavior validates our architectural choice of using a learnable gate rather than fixed fusion weights.

Interestingly, the gate values show significant variance (standard deviation ~0.25), indicating that the model truly adapts its fusion strategy on a per-example basis rather than learning a single global weighting. This instance-specific adaptation is crucial for handling the diverse ways that hate can manifest in multimodal memes.

\subsection{Computational Efficiency}

GatedCLIP maintains computational efficiency despite its improved performance. With only 350K trainable parameters (0.2\% of CLIP's 151M parameters), the model adds minimal overhead. Training completes in approximately 40 minutes on a single GPU for 10 epochs. Inference is fast enough for real-time content moderation, processing validation examples at over 100 examples per second.

This efficiency stems from our design choice to freeze the CLIP encoders and add only lightweight projection and fusion layers. The frozen encoders can be shared across multiple tasks or users, with only the small task-specific heads needing to be stored or transmitted.

\section{Limitations}

Our work has several limitations that suggest directions for future research. First, we evaluate exclusively on the Hateful Memes dataset. While this dataset is specifically designed to test multimodal reasoning, it remains unclear how well GatedCLIP generalizes to other forms of hateful content or different multimodal classification tasks. The specific characteristics of internet memes may differ from other multimodal content like social media posts or video content.

Second, our contrastive loss makes the simplifying assumption that all paired image-text representations should be similar regardless of whether they constitute hateful content. However, the specific features relevant for hate detection might benefit from different alignment patterns than those learned by CLIP's general-purpose pretraining. A more sophisticated contrastive objective tailored specifically for hate detection might further improve performance.

Third, cultural and contextual nuances play a crucial role in determining whether content is hateful. CLIP's pretraining data predominantly consists of English text and Western internet content, which may limit the model's ability to detect hate that relies on culture-specific context or non-English language. Evaluating and adapting our approach for diverse cultural and linguistic contexts remains important future work.

Finally, our validation AUROC of 0.66, while substantially better than the baseline, still leaves significant room for improvement. The Hateful Memes Challenge winner achieved over 0.80 AUROC using more complex architectures and ensemble methods. Future work could explore combining our gated fusion approach with other architectural innovations or investigating why certain examples remain challenging for the model.

\section{Conclusion}

We have presented GatedCLIP, a parameter-efficient framework for detecting hateful content in multimodal memes. By augmenting a frozen CLIP backbone with learned projection heads, a dynamic gated fusion mechanism, and contrastive alignment, we effectively address the limitations of naive feature combination strategies. Our experiments reveal that while simple averaging fails to capture the nuance of hate speech, our proposed architecture aligns and fuses these modalities effectively, achieving 0.66 AUROC. This represents a 35\% relative improvement, demonstrating the critical importance of task-specific adaptation layers.

The success of our approach highlights a key finding: lightweight architectural modifications can unlock the discriminative power of foundation models without the need for expensive end-to-end fine-tuning. While CLIP provides strong general-purpose representations, our results show that the "semantic gap" between modalities in hateful memes requires explicit modeling. The learned gate's adaptive behavior further validates this, as it successfully creates a dynamic balance between visual and textual signals based on the complexity of each instance.

Future work will focus on three key directions. First, we aim to bridge the gap towards fully fine-tuned SOTA models by exploring advanced adapter architectures and prompt learning techniques like CoOp. Second, given the transparent nature of our gating mechanism, we plan to improve interpretability by analyzing how the gate values correlate with specific hate speech categories. Finally, we intend to evaluate the model's robustness in low-resource\textbf{ }languages and few-shot scenarios, extending the applicability of GatedCLIP to diverse cultural contexts beyond the Western internet sphere.

\newpage
\section*{NeurIPS Paper Checklist}

\begin{enumerate}

\item {\bf Claims}
    \item[] Answer: \answerYes{}
    \item[] Justification: The abstract and introduction clearly state our contributions: gated fusion mechanism, projection heads, and contrastive learning for hateful memes detection. Section 5 provides experimental validation showing AUROC improvement from 0.49 to 0.66.

\item {\bf Limitations}
    \item[] Answer: \answerYes{}
    \item[] Justification: Section 7 provides a dedicated discussion of limitations including single dataset evaluation, simplifying assumptions in the contrastive loss, cultural context limitations, and room for performance improvement.

\item {\bf Theory assumptions and proofs}
    \item[] Answer: \answerNA{}
    \item[] Justification: This is an empirical paper focused on architectural design and experimental validation rather than theoretical contributions requiring formal proofs.

\item {\bf Experimental result reproducibility}
    \item[] Answer: \answerYes{}
    \item[] Justification: Section 4 provides comprehensive implementation details including model architecture specifications (projection dimensions, gate mechanism), hyperparameters (learning rate, batch size, optimizer), training procedures (warmup schedule, early stopping), and data augmentation strategies necessary for reproduction.

\item {\bf Open access to data and code}
    \item[] Answer: \answerYes{}
    \item[] Justification: We will release our implementation code with detailed documentation upon paper acceptance. The Hateful Memes dataset used in our experiments is publicly available under CC BY 4.0 license at \url{https://ai.facebook.com/tools/hatefulmemes/}.

\item {\bf Experimental setting/details}
    \item[] Answer: \answerYes{}
    \item[] Justification: Section 4 comprehensively describes all training details including optimizer (AdamW with learning rate $1 \times 10^{-4}$, weight decay 0.01), learning rate schedules (warmup + cosine annealing), batch sizes (32), data splits (8,500 train / 500 val), gradient clipping (max norm 1.0), and augmentation strategies (random horizontal flip only).

\item {\bf Experiment statistical significance}
    \item[] Answer: \answerYes{}
    \item[] Justification: To ensure reproducibility, we employed a fixed random seed for our experiments. Due to computational constraints, we report results from this single deterministic run rather than averaging across multiple random seeds. Future work will include multiple runs with statistical analysis. However, the substantial performance gap between the baseline (0.49 AUROC) and our method (0.66 AUROC) suggests the improvement is meaningful..

\item {\bf Experiments compute resources}
    \item[] Answer: \answerYes{}
    \item[] Justification: Section 6.2 reports that training requires approximately 40 minutes for 10 epochs on a single NVIDIA GPU with mixed precision (FP16). Our model adds only 350K trainable parameters (0.2\%) on top of the frozen CLIP encoders (151M parameters), making it computationally efficient.

\item {\bf Code of ethics}
    \item[] Answer: \answerYes{}
    \item[] Justification: Our research aims to improve automated detection of hateful content, which has clear positive societal implications for online safety and content moderation. We use only publicly available data (Hateful Memes dataset) collected with appropriate ethical considerations by the original dataset creators.

\item {\bf Broader impacts}
    \item[] Answer: \answerYes{}
    \item[] Justification: Our work contributes to content moderation systems that can help reduce the spread of online hate speech, representing a positive societal impact through improved platform safety. We acknowledge in Section 7 that cultural context and language limitations may affect real-world deployment across diverse communities.

\item {\bf Safeguards}
    \item[] Answer: \answerNA{}
    \item[] Justification: Our model performs binary classification (hateful vs. benign) rather than content generation, and does not pose significant risks for misuse. The model is designed as a tool to assist human moderators in content review rather than as an autonomous decision-making system.

\item {\bf Licenses for existing assets}
    \item[] Answer: \answerYes{}
    \item[] Justification: We use CLIP (MIT license, \url{https://github.com/openai/CLIP}) as referenced in Section 3 and the Hateful Memes dataset (CC BY 4.0 license) as referenced in Section 4. Both assets are properly cited with references to the original publications~\cite{radford2021learning,kiela2020hateful}.

\item {\bf New assets}
    \item[] Answer: \answerYes{}
    \item[] Justification: We will release our code implementation with comprehensive documentation including model architecture, training scripts, and evaluation code upon paper acceptance. The code will include clear instructions for reproducing our experimental results and will be released under an open-source license.

\item {\bf Crowdsourcing and research with human subjects}
    \item[] Answer: \answerNA{}
    \item[] Justification: Our work uses only the existing publicly available Hateful Memes dataset and does not involve crowdsourcing or new data collection from human subjects. All data annotation was performed by the original dataset creators.

\item {\bf Institutional review board (IRB) approvals or equivalent for research with human subjects}
    \item[] Answer: \answerNA{}
    \item[] Justification: No human subjects research is involved in this work as we use only existing publicly available datasets. All experiments are conducted on pre-collected and pre-annotated data.

\item {\bf Declaration of LLM usage}
    \item[] Answer: \answerNA{}
    \item[] Justification: Large language models were not used as a component of the core methodology presented in this research. Our method is based on vision-language models (CLIP) for multimodal classification, not language generation.

\end{enumerate}

\end{document}